\documentclass[10pt, a4paper]{article}
\usepackage{lrec2022} 
\usepackage{multibib}
\newcites{languageresource}{Language Resources}
\usepackage{graphicx}
\usepackage{tabularx}
\usepackage{soul}
\usepackage{booktabs, multirow} 
\usepackage{makecell}
\usepackage{comment}

\usepackage{titlesec}
\titleformat{\section}{\normalfont\large\bfseries\center}{\thesection.}{1em}{}
\titleformat{\subsection}{\normalfont\SmallTitleFont\bfseries\raggedright}{\thesubsection.}{1em}{}
\titleformat{\subsubsection}{\normalfont\normalsize\bfseries\raggedright}{\thesubsubsection.}{1em}{}
\renewcommand\thesection{\arabic{section}}
\renewcommand\thesubsection{\thesection.\arabic{subsection}}
\renewcommand\thesubsubsection{\thesubsection.\arabic{subsubsection}}

\usepackage{epstopdf}
\usepackage[utf8]{inputenc}

\PassOptionsToPackage{hyphens}{url}\usepackage{hyperref}
\usepackage{xstring}

\usepackage{color}

\title{Sequence-to-Sequence Resources for Catalan}

\name{Ona de Gibert, Ksenia Kharitonova, Blanca Calvo Figueras,  \\ {\bf \large Jordi Armengol-Estapé, Maite Melero}}

\address{Barcelona Supercomputing Center\\
        Plaça Eusebi Güell 1-3, Barcelona 08034, Spain\\
         \{ona.degibert, ksenia.kharitonova, blanca.calvo, jordi.armengol, maite.melero\}@bsc.es\\}

\abstract{
In this work, we introduce sequence-to-sequence language resources for Catalan, a moderately under-resourced language, towards two tasks, namely: Summarization and Machine Translation (MT). We present two new abstractive summarization datasets in the domain of newswire. We also introduce a parallel Catalan$\leftrightarrow$English corpus, paired with three different brand new test sets. Finally, we evaluate the data presented with competing state of the art models, and we develop baselines for these tasks using a newly created Catalan BART. We release the resulting resources of this work under open license to encourage the development of language technology in Catalan.
 \\ \newline  }

\begin{document}

\maketitleabstract

\section{Introduction\label{sec:intro}}
In recent years, the arrival of the transformers \cite{DBLP:journals/corr/VaswaniSPUJGKP17} has changed the landscape of Natural Language Processing (NLP). The potential of this new technology has opened up new lines of research with a clear focus on under-resourced languages \cite{zoph-etal-2016-transfer}, since it has been shown that pre-trained language models such as BERT \cite{devlin_bert_2019} can successfully solve downstream tasks with much less data than what was needed before.

For this reason, academia has made an effort on developing language-specific pre-trained language models \cite{martin2020camembert} and evaluation benchmarks \cite{canete2020spanish}. While this field is still dominated by big languages, we can see how this is starting to change \cite{wang2020extending}.

We focus on Catalan, a moderately under-resourced language, for which there already exists a large monolingual corpus, a pre-trained encoder-only language model, and a Natural Language Understanding (NLU) benchmark \cite{armengol2021multilingual}. We explore the task of Natural Language Generation (NLG) by developing resources for two Sequence-to-Sequence tasks, namely, Summarization and Machine Translation (MT). These are complex tasks that require an encoder-decoder architecture. 

In this paper, we present new resources and models, diverse in sizes and domains. The public release of these resources will allow the Natural Language Processing (NLP) community to explore their algorithms in depth, as well as the cross-lingual transfer-learning capabilities of their models. Our contributions sum up to:
\begin{itemize}
    \item Two new datasets for abstractive Summarization
    \item A high-quality dataset for English$\leftrightarrow$Catalan MT
    \item Three new MT testsets for the English$\leftrightarrow$Catalan pair, one of which is multilingual
    \item A Catalan BART
\end{itemize}
We make these resources openly available at Github.\footnote{\url{https://github.com/TeMU-BSC/seq-to-seq-catalan}}

The rest of the paper is organized as follows. Section \ref{sec:related} provides an overview of the previous work done in the field. Section \ref{sec:language_resources} describes in detail the resources presented. Section \ref{sec:experiments} describes the experiments and results obtained, and finally, section \ref{sec:conclusions} concludes our work and opens new future lines of research.

\section{Related Work\label{sec:related}}



Language resources for summarization are difficult to obtain and thus are often built with automated processes from news websites, as it is easy to interpret the body of an article and its title. There exist several resources for English, both extractive \cite{lins2019cnn} and abstractive \cite{narayan2018don} and recently there has been an effort on developing multilingual resources \cite{scialom2020mlsum}, with a focus on big languages. For the case of Catalan, a minority language, there exists only one resource provided by \newcite{ahuir2021nasca}, who collected newspaper articles from different sources for Spanish and Catalan and trained a Transformer encoder-decoder for both languages. For Catalan they created DACSA, a text summarization dataset composed of 725,184 sample pairs from 9 newspapers. They benchmark their results with two well known multilingual models: mBART \cite{liu2020multilingual} and mT5 \cite{xue2021mt5} to investigate whether monolingual encoder-decoders are more beneficial than multilingual ones.

In the case of MT, there has been much work on developing resources for under-resourced languages. One of the main repositories of parallel data for MT is OPUS \citelanguageresource{tiedemann2012parallel}, which includes many multilingual parallel datasets ranging in domains, languages and sizes. Catalan is included in many of these large web-crawled datasets, however, as \newcite{kreutzer2021quality} point out, most data coming from online sources is of poor quality. Hence, the importance of high quality data curation. In this work, we focus on the language pair English-Catalan, for which, even if English is a \textit{lingua franca}, there are not many publicly available parallel resources. These are mostly from OPUS or from Softcatala,\footnote{\url{softcatala.org}} an non-profit organization that works for the development of technologies in Catalan.

Over the past few years, large Transformer-based \cite{NIPS2017_3f5ee243} models have shown to yield the best results on the majority of the sequence-to-sequence tasks. The most common approach towards developing such models has been to pre-train them on a vast amount of rich non-parallel text data with a variety of objectives \cite{devlin_bert_2019}, and afterwards fine-tune them on a small amount of appropriate data for a required downstream task. This approach can be used both with monolingual corpora for primarily monolingual tasks, and with multilingual data. Following this approach, denoising autoencoders expanded on an original sequence-to-sequence Transformer and they proved to be especially useful for summarization due to their denoising objective \cite{lewis2019bart}. Multilingual BART (mBART), pre-trained on a multilingual concatenated non-parallel data, further showed performance gain in a low resource language setting for machine translation \cite{liu2020multilingual}.  
 

\section{Language Resources\label{sec:language_resources}}
\subsection{Summarization\label{sec:sum_data}}

We introduce two new datasets for summarization in Catalan. CaSum, which can be used for training and evaluation, and VilaSum, an out-of-distribution (with respect to CaSum) test set.

\paragraph{CaSum}  is a new summarization dataset. It is extracted from a newswire corpus crawled from the Catalan News Agency.\footnote{\url{https://www.acn.cat/}}

The corpus consists of 217,735 instances that are composed by the headline and the body. We obtained each headline and its corresponding body and applied the following cleaning pipeline: deduplicating the documents, removing the documents with empty attributes, and deleting some boilerplate sentences. 

Since most summarization datasets are built automatically, there's little control over the final quality result. We perform a manual evaluation of the dataset by taking 10,000 random samples and assessing whether the headline is a valid summary for the article. The validation reports that 99,02 \% of the cases are correct. Therefore, we consider our dataset of high quality.

\paragraph{VilaSum} is a smaller summarization dataset, of 13,843 samples, which can be used as an out-of-distribution (with respect to CaSum) test set. It has been obtained from the digital newspaper Vilaweb.\footnote{\url{https://www.vilaweb.cat/}} To obtain the final dataset, we followed the same procedure as in the CaSum dataset. However, because of its smaller size, we were able to manually validate all the samples of the dataset. The actual crawling returned 15,019 headline-body pairs and the manual revision discarded 1,176 pairs, where the headline was not a valid summary of the article.

The splits for both datasets can be seen in Table \ref{tab:sum}. Table \ref{tab:sum_stats} better describes each dataset in terms of article and summary length, as well as novelty (how many words in the summary do not belong to the article), compression ratio, and vocabulary size. We note that VilaSum has longer articles, with shorter summaries and more novel words in the summaries, making this test more challenging than the one in CaSum. 

\begin{table}[t]
\centering
\begin{tabular}{l|ccc}
Dataset & Train & Valid & Test\\
\hline
CaSum & 197,735 & 10,000 & 10,000  \\
VilaSum & - & - &  13,843  \\
\end{tabular}
\caption{\label{tab:sum}Splits of the Summarization datasets}
\end{table}

\begin{table}[t]
\centering
\begin{tabular}{l|rr}
& CaSum & VilaSum \\
\hline
Article avg. sentences &11.69 &29.34 \\
Summary avg. sentences &1.00 &1.01 \\
\hline
Article avg. words &338.64 &647.69 \\
Summary avg. words &16.42 &12.89 \\
\hline
Novelty &19.66 &25.75 \\
Compression ratio &21.25 &57.7 \\
Vocabulary Size &1,052,211 &380,146 \\
\end{tabular}
\caption{\label{tab:sum_stats} Statistics of the Summarization datasets}
\end{table}



\subsection{Machine Translation\label{sec:mt_data}}

In order to create a large dataset for CA$\leftrightarrow$EN machine translation, we compile all available open-source parallel bilingual CA$\leftrightarrow$EN corpora, plus a brand new high quality dataset, gEnCaTa. In total, we use 20 different datasets to obtain a moderately large bilingual corpus CA$\leftrightarrow$EN resulting in a dataset of over 11.3 million aligned sentences. We release openly the gEnCaTa corpus. The characteristics of the corpora can be found in Table \ref{tab:mt-data}. 

\begin{table*}[t]
\centering
\begin{tabular}{llrll}
ID & Dataset & Sentences & Source & Domain \\
\hline
1 &CCaligned &5,787,682 & \citelanguageresource{elkishky_ccaligned_2020} & General \\
2 & COVID-19 Wikipedia &1,531 & \citelanguageresource{tiedemann2012parallel} & Health \\
3 &CoVost en-ca &79,633 & \citelanguageresource{wang2020covost} & General\\
4 &CoVost ca-en &263,891 & \citelanguageresource{wang2020covost} & General \\
5 &Eubookshop &3,746 &  \citelanguageresource{tiedemann2012parallel} & Legislation \\
6 &Europarl &1,965,734 & \citelanguageresource{tiedemann2012parallel} & Legislation \\
7 &Global Voices &21,342 & \citelanguageresource{tiedemann2012parallel} & Newswire\\
8 &Gnome &2,183& \citelanguageresource{tiedemann2012parallel} & Software\\
9 &JW300 &97,081 & \citelanguageresource{tiedemann2012parallel} & General \\
10 &KDE4 &144,153 & \citelanguageresource{tiedemann2012parallel} & Software\\
11 &Memories Lliures &1,173,055 & Softcatalà & General \\
12 &Open Subtitles &427,913 & \citelanguageresource{lison-tiedemann-2016-opensubtitles2016} & General\\
13 &Books &4,580 &\citelanguageresource{tiedemann2012parallel} & Narrative \\
14 &QED &69,823 &\citelanguageresource{abdelali-etal-2014-amara} & Education \\
15 &Tatoeba &5,500 &\citelanguageresource{tiedemann2012parallel} & General \\
16 &Tedtalks &50,979 & Softcatalà  & General\\
17 &Ubuntu &6,781  &\citelanguageresource{tiedemann2012parallel} & Software\\
18 &Wikimatrix &977,466 & \citelanguageresource{schwenk2019wikimatrix} & Wikipedia \\
19 &Wikimedia &208,073 &\citelanguageresource{tiedemann2012parallel} & Wikipedia \\
20 &GEnCaTa &38,595& New & General \\\hline
&Total &11,329,741 \\
\end{tabular}
\caption{\label{tab:mt-data}Language resources for Machine Translation Training.}
\end{table*}

The compiled datasets originate from different sources and belong to different domains. Mostly they come from OPUS \cite{tiedemann2012parallel} and Softcatalà.\footnote{\url{https://github.com/Softcatala/parallel-catalan-corpus}} Most datasets belong to the general domain, although some sources originate from software translations or Wikipedia articles. Nonetheless, we are aware that the quality of the datasets varies greatly, since an automatic alignment and manual revision yield very different results. CCaligned, for instance, has been shown to have poor quality \cite{kreutzer2021quality}.


\paragraph{gEnCaTa} is a Catalan$\leftrightarrow$English parallel corpus composed of 38,595 segments.
It has been compiled by leveraging parallel data from crawling the \url{gencat.cat} domain and subdomains, belonging to the Catalan Government, both in English and Catalan.

We used the cleaning pipeline in \newcite{armengol-estape-etal-2021-multilingual} pipeline to process the WARC files obtained from the crawling. This allowed us to maintain the metadata and retrieve the original URL per each visited page. We extracted the content of the URLs that had data crawled in both languages and obtained 4,416 comparable sites, which we consider documents. 

To align the sentences in the documents we used both an automated alignment algorithm (vecalign\footnote{\url{https://github.com/thompsonb/vecalign}}), and a manual revision. After the automated alignment and sentence deduplication there were 51,447 parallel segments. However, after the manual revision only 38,595 remained. We did a further analysis of these differences just to notice that only 19.8\% of the 5,000 highest scored segments ranked by vecalign were also selected after the manual revision. This posits the question of how much can we rely on alignment algorithms for building parallel corpora by only looking at the given score.



\subsubsection{Machine Translation Evaluation\label{sec:mt_data_eval}}

We also provide three new datasets for MT evaluation.

\paragraph{WMT2013-ca} consists of the Catalan translation of the WMT 2013 translation shared task test set \cite{bojar-etal-2013-findings}, belonging to the newswire domain. We commissioned the translation from Spanish to Catalan to a professional native translator.

\paragraph{taCon} is a multilingual test set from the legal domain that includes all the languages in which the Spanish Constitution exists, namely, Basque, Catalan, Galician, Spanish and English.
To obtain it, we downloaded the Spanish Constitution from the website of the Agencia Estatal del Boletín Oficial del Estado\footnote{\url{www.boe.es}} in the corresponding languages in PDF format. We converted it to plain text, fixed the broken sentences and finally aligned the sentences manually.

\paragraph{Cyber-ca} is a brand new test set in Catalan, Spanish and English that belongs to the cybersecurity domain. It is composed of cybersecurity alerts extracted from the INCIBE Spanish-English corpus.\footnote{\url{https://www.elrc-share.eu/repository/browse/descripciones-de-vulnerabilidades-de-la-bbdd-nvd/9dd01c02cfcc11e9913100155d02670670070560807f458f8e50bd5273778d7a/}}

\begin{table}[t]
\centering
\begin{tabular}{llll}
Dataset & Languages & Domain & Sent.\\\hline
WMT13-ca  & ca, es, en & newswire & 3,003 \\
Cyber-ca & ca, en, es & cybersecurity & 1,715 \\
taCon & \makecell{ ga, eu, ca, \\ es, en }   & legal & 1,110 \\
\end{tabular}
\caption{\label{tab:mt-tests}Language resources for Machine Translation Evaluation}
\end{table}

\section{Experiments and Results\label{sec:experiments}}

\subsection{Summarization\label{sec:sum_exp}}

We develop a Catalan BART-base \cite{lewis2019bart}, using the Catalan Textual Corpus \cite{armengol2021multilingual}, henceforth BART-Ca. We fine-tune it with the CaSum training and validation sets during circa 4 epochs, using performance during the validation as a stop signal. We evaluate our model on the CaSum test set and also on VilaSum. We benchmark our results with mBART \cite{liu2020multilingual}, also fine-tuning it with CaSum during approximately the same amount of epochs; and with NASCA,\footnote{Due to 512 token size limit, we can only evaluate 6,742 and 7,199 samples of each testset for CaSum and VilaSum, respectively.} the pre-trained Catalan language model by \newcite{ahuir2021nasca}. We report our results with ROUGE \cite{lin2004rouge} in Table \ref{tab:sum_results}.

\begin{table}[!htp]\centering
\begin{tabular}{llccc}
Test set &Model &ROUGE-1 & ROUGE-L  \\
\hline
\multirow{3}{*}{CaSum} &BART-Ca & 41.39 & 36.14 \\
&NASCA & 24.42 & 19.89  \\
&mBART & 43.95 & 38.11 \\
\hline
\multirow{3}{*}{VilaSum} &BART-Ca & 35.04 & 29.70 \\
&NASCA & 23.18& 19.09 \\
&mBART & 33.17 & 27.52 \\
\end{tabular}
\caption{Summarization results on CaSum and VilaSum}\label{tab:sum_results}
\end{table}


BART-Ca obtains the best results on the VilaSum dataset, whereas mBART performs better for the CaSum test set itself. Therefore, BART-Ca shows better performance on the out-of-distribution data, despite being considerably smaller.\footnote{BART-base vs. BART large, that is, 110M vs. 340M non-embedding parameters.} This may be attributed to the fact that BART-Ca is trained on a more optimized vocabulary (language-specific), and pre-trained on data of better quality. Results show that the NASCA model significantly underperforms both fine-tuned BART-Ca and mBART, both on the test CaSum and the VilaSum datasets. This can be expected, as NASCA has not been fine-tuned, and the distribution of the data it was trained on may differ from our test sets.

With regard to good performance on the test CaSum dataset, we can assume that this dataset is just a continuation of the training and validation datasets that were used to fine-tune the models and, therefore, is exactly in the same domain and in the same style. However, the performance on the VilaSum dataset shows that the quality of the data used to fine-tune the starting pre-trained model is of tantamount importance. 

\subsection{Machine Translation\label{sec:mt_exp}}

To test the new evaluation sets, we use two supervised neural MT models: Google Translate (GT) \footnote{\url{https://translate.google.es}} and Softcatalà,\footnote{\url{https://www.softcatala.org/traductor/}} the latter one is open-source and has established itself as a reference in the Catalan community. As a reference, we also include the test set of Flores-101 \citelanguageresource{goyal2021flores}, a multilingual test set for MT benchmarking. We report BLEU scores \cite{post-2018-call} in Table \ref{tab:bleu}.

\begin{table}[!htp]\centering
\begin{tabular}{llccc}
Direction &Test set & GT &Softcatalà \\\hline
\multirow{4}{*}{en $\rightarrow$ ca} &WMT13 &33.8 &34.3 \\
&TaCon  & 37.6 & 31.8\\
&Cyber &46.5 &42.1 \\
&Flores-101 &42.2 &41.5 \\\hline
\multirow{4}{*}{ca $\rightarrow$ en} &WMT13  &39.8 &37.6 \\
&TaCon & 43.2 & 35.2\\
&Cyber  &58 &49.9 \\
&Flores-101 &46.9 &42.4 \\
\end{tabular}
\caption{BLEU scores for MT evaluation}
\label{tab:bleu}
\end{table}

Results show that MT achieves overall good results for the CA$\leftrightarrow$EN language pair, with more mediocre results in certain evaluation sets. As expected, GT outperforms Softcatalà, but the latter is still a competitive baseline. 

TaCon is the test set with the lowest scores, probably because of its restricted language-specific domain. Surprisingly, the Cyber test set seems to be the easiest one, although it is domain-specific. This is because it contains numerous non-verbal segments that are kept untranslated, which boosts the results, achieving 58 BLEU for GT in the CA$\rightarrow$EN direction. Both Flores-101 and WMT13, which belong to the more general domain, present almost no variability in the models' performance.

\section{Conclusions \& Future Work\label{sec:conclusions}}

In this work we presented resources for two sequence-to-sequence tasks, specifically, Summarization and Machine Translation, and for a moderately under-resourced language, Catalan. We describe in detail two new high-quality Summarization datasets and several resources to exploit MT in Catalan. We further explore our results by building baselines and comparing them with state of the art models. We expect to encourage the development of more complex language technologies for this language. As future work, we plan to develop new training and evaluation resources for Catalan, noting that the generative scenario is the one that currently lacks data.

\section{Acknowledgements}
This work was funded by the MT4All CEF project.\footnote{\url{https://ec.europa.eu/inea/en/connecting-europe-facility/cef-telecom/2019-eu-ia-0031}}
\section{Bibliographical References}\label{reference}

\bibliographystyle{lrec2022-bib}
\bibliography{lrec2022-example}

\section{Language Resource References}
\label{lr:ref}
\bibliographystylelanguageresource{lrec2022-bib}
\bibliographylanguageresource{languageresource}

\end{document}